\documentclass[11pt,a4paper]{article}
\usepackage[hyperref]{emnlp2020}
\usepackage{times}
\usepackage{latexsym}

\usepackage{graphicx}
\usepackage{subfigure}
\usepackage{enumitem}
\usepackage{multirow}
\usepackage{booktabs}
\usepackage{url}
\usepackage{microtype}

\aclfinalcopy 

\title{GraphDialog: Integrating Graph Knowledge into End-to-End Task-Oriented Dialogue Systems}

\author{Shiquan Yang, Rui Zhang\Thanks{ Rui Zhang is the corresponding author.}, Sarah Erfani\\
The University of Melbourne, Australia\\
\{shiquan@student., rui.zhang@, sarah.erfani@\}unimelb.edu.au}

\date{}

\begin{document}
\maketitle
\begin{abstract}
End-to-end task-oriented dialogue systems aim to generate system responses directly from plain text inputs. There are two challenges for such systems: one is how to effectively incorporate external knowledge bases (KBs) into the learning framework; the other is how to accurately capture the semantics of dialogue history. In this paper, we address these two challenges by exploiting the graph structural information in the knowledge base and in the dependency parsing tree of the dialogue. To effectively leverage the structural information in dialogue history, we propose a new recurrent cell architecture which allows representation learning on graphs. To exploit the relations between entities in KBs, the model combines multi-hop reasoning ability based on the graph structure. Experimental results show that the proposed model achieves consistent improvement over state-of-the-art models on two different task-oriented dialogue datasets.

\end{abstract}

\section{Introduction}
\begin{figure}[t]
\setlength{\belowcaptionskip}{-3mm}
    \centering
    \includegraphics[width=3.0in]{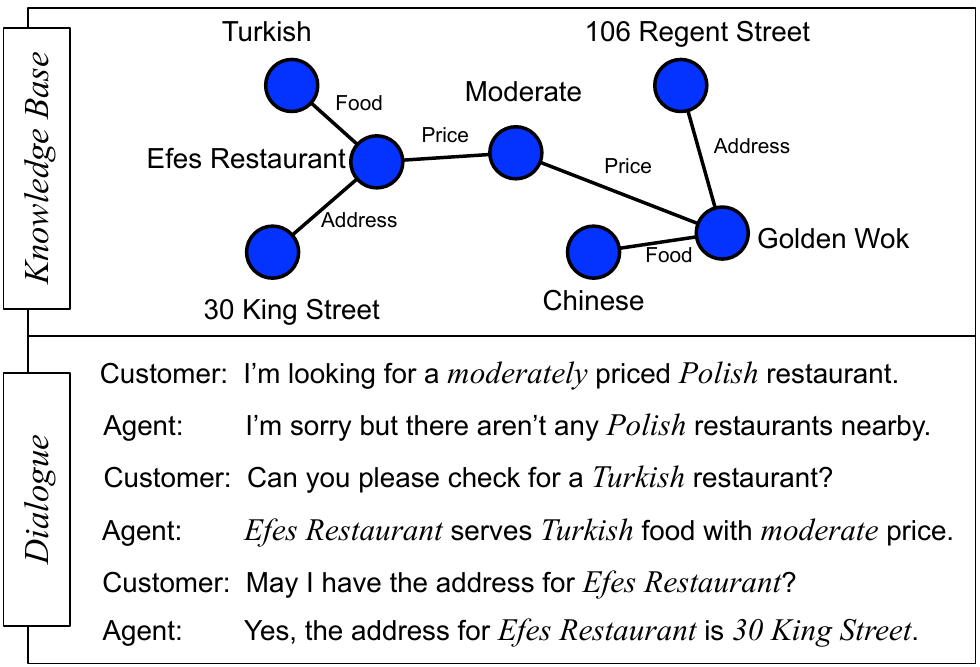}
    \caption{An example dialogue in the restaurant booking domain. The top part is knowledge base (KB) information that represented by a graph and the bottom part is the conversation between a customer and the agent. Our aim is to predict the agent responses given KB information and the customer utterances.}
    \label{dialogue_example}
\end{figure}

Task-oriented dialogue systems aim to help user accomplish specific tasks via natural language interfaces such as restaurant reservation, hotel booking and weather forecast. There are many commercial applications of this kind (e.g. Amazon Alexa, Google Home, and Apple Siri) which make our life more convenient. Figure \ref{dialogue_example} illustrates such an example where a customer is asking for the information about restaurants. By querying the knowledge base (KB), the agent aims to provide the correct restaurant entities from the KB to satisfy the customer in a natural language form. Hence, the ability to understand the dialogue history, and to retrieve relevant information from the KB is essential in task-oriented dialogue systems.

One approach for designing task-oriented dialogue systems is the pipeline approach \citep{williams2007partially,lee2009example,young2013pomdp}, but it suffers from the difficulty in credit assignment and adaption to new domains. Another popular approach is the end-to-end models \citep{serban2016building,wen2016network,williams2017hybrid,zhao2017generative,serban2017hierarchical}, which directly map the dialogue history to the output responses. This approach has attracted more attention in the research community recently as it alleviates the drawbacks of the pipeline approach. 
However, end-to-end dialogue models usually suffer from ineffective use of knowledge bases due to the lack of appropriate framework to handle KB data.

To mitigate this issue, recent end-to-end dialogue studies \citep{eric2017key,mem2seq} employ memory networks \citep{weston2014memory,sukhbaatar2015end} to support the learning over KB, and have achieved promising results via integrating memory with copy mechanisms \citep{gulcehre2016pointingunknownwords,eric2017copyaugmented}. By using memory, they assume that the underlying structure of KB is linear since memory can be viewed as a list structure. As a result, the relationships between entities are not captured. However, since KB is naturally a graph structure (nodes are entities and edges are relations between entities). By overlooking such relationships, the model fails to capture substantial information embedded in the KB including the semantics of the entities which may significantly impact the accuracy of results. Moreover, structural knowledge such as dependency relationships has recently been investigated on some tasks (e.g., relation extraction) \citep{peng2017cross,song2018n} and shown to be effective in the model's generalizability. However, such dependency relationships (essentially also graph structure) have not been explored in dialogue systems, again missing great potential for improvements.

With the above insight, we propose a novel \textbf{graph}-based end-to-end task-oriented \textbf{dialog}ue model (\textbf{GraphDialog}) aimed to exploit the graph knowledge both in dialogue history and KBs. Unlike traditional RNNs such as LSTM \citep{hochreiter1997long} and GRU \citep{cho2014learning}, we design a novel recurrent unit (Section \ref{graphencodercell}) that allows multiple hidden states as inputs at each timestep such that the dialogue history can be encoded with graph structural information. The recurrent unit employs a masked attention mechanism to enable variable input hidden states at each timestep. Moreover, We incorporate a graph structure (Section \ref{graphknowledge}) to handle the external KB information and perform multi-hop reasoning on the graph to retrieve KB entities.

Overall, the contributions of this paper are summarized as follows:
\begin{itemize}
    \item We propose a novel graph-based end-to-end dialogue model for effectively incorporating the external knowledge bases into task-oriented dialogue systems.
    \item We further propose a novel recurrent cell architecture to exploit the graph structural information in the dialogue history. We also combine the multi-hop reasoning ability with graph to exploit the relationships between entities in the KB.
    \item We evaluate the proposed model on two real-world task-oriented dialogue datasets (i.e., SMD and MultiWOZ 2.1). The results show that our model outperforms the state-of-the-art models consistently.
\end{itemize}

\begin{figure*}
\setlength{\belowcaptionskip}{-1mm}
    \centering
    \includegraphics[width=6.3in]{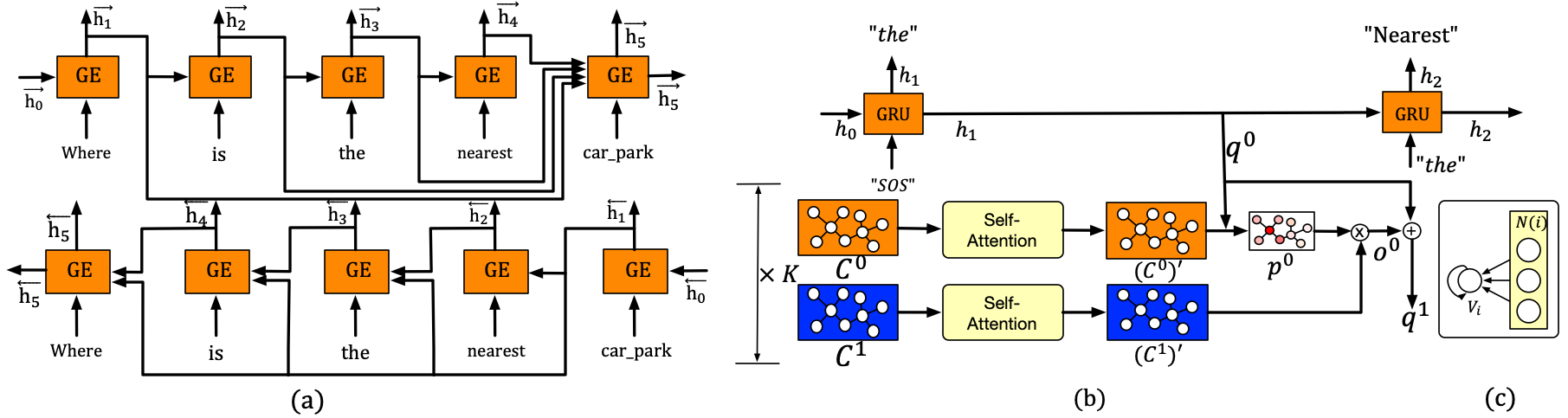}
    \caption{Overview of the proposed architecture. (a) Graph Encoder, top is forward graph and bottom is backward graph. (b) Decoder and Knowledge Graph with multi-hop reasoning mechanism. (c) Self-Attention Mechanism.}
    \label{overallframework}
\end{figure*}

\section{Related Work}
Task-oriented dialogue system has been a long-standing studied topic \citep{williams2007partially,lee2009example,huang2020semi} and can be integrated into many practical applications such as virtual assistant \citep{sun2016contextual,sun2017collaborative}. Traditionally, task-oriented dialogue systems are built in the pipeline approach, which consists of four essential components: natural language understanding \citep{chen2016end}, dialogue state tracking \citep{lee2016task,zhong2018global,wu2019transferable}, policy learning \citep{su2016line,peng2018deep,su2018discriminative} and natural language generation \citep{sharma2016natural,chen2019semantically,huang2019mala}. Another recent approach is the end-to-end models \citep{wu2018end,lei2018sequicity}, which directly map the user utterances to responses without heavy annotations. \citet{bordes2016endtoendtaskorientedlearning} apply end-to-end memory networks \citep{sukhbaatar2015end} for task-oriented dialogues and shown that end-to-end models are promising on the tasks. To produce more flexible responses, several generative models are proposed \citep{zhao2017generative,serban2016building}. They formulate the response generation problem as a translation task and apply sequence-to-sequence (Seq2Seq) models to generate responses. Seq2Seq models have shown to be effective in language modeling but they struggle to incorporate external KB into responses. To mitigate this issue, \citet{eric2017copyaugmented} has enhanced the Seq2Seq model by adding copy mechanism. \citet{mem2seq} combines the idea of pointer with memory networks and obtained improved performance. \citet{wu2019global} incorporates global pointer mechanism and achieved improved performance. Our study differs from those works in that we exploit the powerful graph information both contained in the dialogue history and in the KBs to effectively incorporate KBs into dialogue systems.

\section{Proposed Model}
Our proposed model consists of three components: an encoder (Section \ref{graphencoder}), a decoder (Section \ref{graphdecoder}) and a knowledge graph with multi-hop reasoning ability (Section \ref{graphknowledge}). Formally, let \textit{X} = \{$x_1$,...,$x_n$\} be a sequence of tokens, where each token $x_i\in$ \textit{X} corresponds to a word in the dialogue history. We first obtain a \textit{dialogue graph} \textit{$\widehat{G}$} (Section \ref{dialoguegraph}), which is the dependency parsing graph of the sentences in the dialogue history \textit{X}, as the input of the encoder. The encoder then learns a fixed-length vector as the encoding of the dialogue history based on \textit{$\widehat{G}$}, which is then fed to the decoder for hidden state initialization. The knowledge graph adopts another graph \textit{G} = \{\textit{V},\textit{E}\} to store and retrieve the external knowledge data (Section \ref{graphcontent}), where \textit{V} denotes the entities and \textit{E} denotes the edges. The decoder generates the system response \textit{Y} = \{$y_1$,...,$y_m$\} token-by-token either by copying entities from graph \textit{G} via querying the knowledge graph or by generating tokens from vocabularies. Figure \ref{overallframework} illustrates the overall architecture of the proposed model. In the following sections, we describe each component in detail.

\subsection{Graph Encoder}\label{graphencoder}

\subsubsection{Dialogue Graph}\label{dialoguegraph}
\begin{figure}[t]
    \centering
    \includegraphics[width=3.0in]{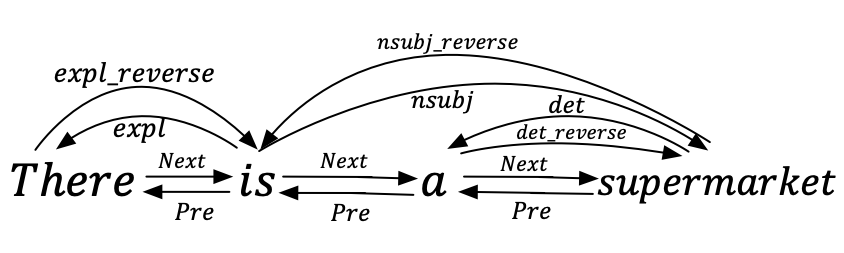}
    \caption{An example of dialogue graph.}
    \label{dependencyparsingresult}
\end{figure}

To enable learning semantic rich representations of words with various relationships, such as adjacency and dependency relations, we first use the off-the-shelf tool \textit{spacy}\footnote{\url{https://spacy.io/}} to extract the dependency relations among the words in the dialogue history \textit{X}. Figure \ref{dependencyparsingresult} gives an example of the dependency parsing result. The bi-directional edges among words allow information flow both from dependents to heads and from heads to dependents. The intuition is that the representation learning of the head words should be allowed being influenced by the dependent words and vice versa, thus allowing the learning process to capture the mutual relationships between the head words and the dependent words to provide richer representation.

We compose the dialogue graph by combining the obtained dependency relations with the sequential relations (i.e., \textit{Next} and \textit{Pre}) among words, which serves as the input to the graph encoder. To further support bi-directional representation learning, we split the obtained dialogue graph into two parts: the \textit{forward graph} (from left to right) and the \textit{backward graph} (from right to left).

\subsubsection{Recurrent Cell Architecture}\label{graphencodercell}
\begin{figure}[t]
\setlength{\belowcaptionskip}{-1mm}
    \centering
    \includegraphics[width=2.5in]{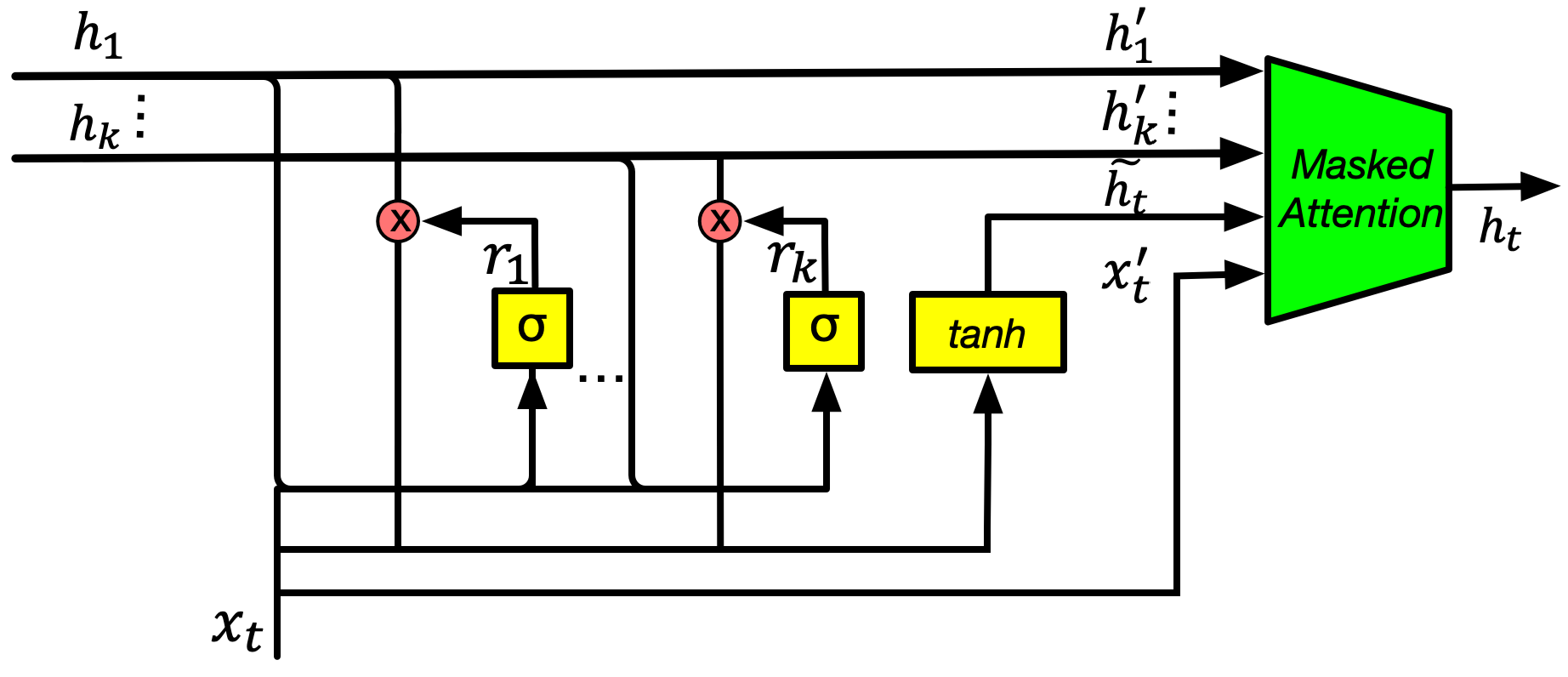}
    \caption{Overview of the proposed recurrent unit.}
    \label{recurrentcell}
\end{figure}

The recurrent cell architecture (Figure \ref{recurrentcell}) is the core computing unit of the graph encoder, and is used to compute the hidden state of each word in the obtained dialogue graph. The cell traverse the words in the dialogue graph sequentially according to the word order in the dialogue history. Next, we show how to compute the cell hidden state \textit{$h_t$} at timestep \textit{t}.

 Let us define \textit{$x_t$} as the input word representation at timestep \textit{t}. \textit{P(t)} = \{$p_1$,$p_2$,\dots,$p_k$\} is the set of precedent words for \textit{$x_t$} where each $p_i\in$ \textit{P(t)} denotes a word in the dialogue graph that connects to \textit{$x_t$}, and \textit{k} is the total number of the precedents of \textit{$x_t$}. \textit{H} = \{$h_1$,$h_2$,\dots,$h_k$\} is a set of hidden states where each element $h_j\in$ \textit{H} denotes the hidden state of the \textit{j}-th predecessor $p_j\in$ \textit{P(t)}.
 
The input of the cell consists of two parts: the input word vector \textit{$x_t$}, and the predecessor hidden states \textit{H}. First, we loop over the \textit{k} hidden states in \textit{H} and compute a \textit{reset gate} for each of them. Specifically, we compute \textit{$r_j$} for the \textit{j}-th hidden state using:

\vspace{-2mm}
 \begin{equation}
     \textit{$r_j$} = \sigma\left(W_r x_t + U_r h_j \right) \label{resetgate}
 \end{equation}

\noindent where $\sigma$ is the logistic sigmoid function, \textit{$x_t$} and \textit{$h_j$} are the current input and the hidden state of the \textit{j}-th predecessor at timestep \textit{t} respectively. \textit{$W_r$} and \textit{$U_r$} are parameters which will be learned. We then compute a candidate hidden state $\widetilde{h_t}$  using:

\vspace{-3mm}
\begin{equation}
    \widetilde{h_t} = \phi\left(W_n x_t + \frac{1}{k} \sum_{j=1}^{k} r_j * \left( U_n h_j \right) \right)
\end{equation}

\noindent where $\phi$ is the hyperbolic tangent function, \textit{k} is the number of predecessors of word \textit{$x_t$}, \textit{$W_n$} and \textit{$U_n$} are the learnable weight matrices. Intuitively, $\widetilde{h_t}$ is the contextualized representation of current input \textit{$x_t$}.

Next, we combine the obtained candidate hidden state $\widetilde{h_t}$ with the predecessor hidden states \textit{H}, and use an \textit{masked attention mechanism} (Equation \ref{maskedattentionequation}) to aggregate them together to yield the output hidden state \textit{$h_t$} at timestep \textit{t}. To obtain sufficient expressive power, we first apply linear transformations to the input \textit{$x_t$} and the hidden states \textit{$h_j\in$} \textit{H} using:

\vspace{-3mm}
\begin{equation}
    \textit{$x_t^{'}$} = W_z x_t
\end{equation}

\vspace{-3mm}
\begin{equation}
    \textit{$h_j^{'}$} = U_z h_j 
\end{equation}

\noindent where \textit{$W_z$}, \textit{$U_z$} are parameters which are learned, \textit{t} is the current timestep. We denote \textit{$H^{'}$}=\{\textit{$h_1^{'}$},\textit{$h_2^{'}$},\dots,\textit{$h_k^{'}$}\} as the transformed set of hidden states. Then we add the previously obtained candidate hidden state $\widetilde{h_t}$ into the transformed set of hidden states \textit{$H^{'}$} and obtain \textit{$H^{''}$}=\{$h_1^{'}$,$h_2^{'}$,\dots,$h_k^{'}$,$\widetilde{h_t}$\}. The intuition is that the output hidden state depends on both the history information ($h_1^{'}$ to $h_k^{'}$) and the current input ($\widetilde{h_t}$).

Then we perform attention mechanism by using the hidden states \textit{$H^{''}$} as keys and the current input \textit{$x_t$} as query. Intuitively, different inputs (e.g. different predecessors in \textit{$H^{''}$}) should have different impacts on the output hidden state \textit{$h_t$}, and we expect our model to capture that. However, the inputs may have different number of predecessors at different timesteps. To handle this, inspired by \citep{vaswani2017attention}, we employ an \textit{masked attention mechanism} to learn the importance of each predecessor at every timestep, thus avoiding the pad information affecting the learning process. We compute the attention using:

\vspace{-1mm}
\begin{equation}
    \textit{$e_j$} = \textit{$v^{T}$} \phi\left(x_t^{'} + h_j^{'} \right)
\end{equation}

\vspace{-1mm}
\begin{equation}
    \alpha_{j} = \textit{softmax} \left( [e_j]_{m} \right) \label{maskedattentionequation}
\end{equation}

\noindent where \textit{$v$} is a learnable parameter, \textit{$h_j^{'}$} is the \textit{j}-th vector in \textit{$H^{''}$}, \textit{$softmax$}(\textit{$z_i$})={\textit{$e^{z_i}$}}/{$\sum_{j}e^{z_j}$}, $\alpha_{j}$ denotes the attention weight on the \textit{j}-th vector in \textit{$H^{''}$}, $\left[\cdot\right]_{m}$ denotes the mask operation. In our implementation, we simply set the number to negative infinity if the \textit{j}-th hidden state corresponds to a pad token. Finally, we compute the weighted sum to obtain the cell output hidden state \textit{$h_t$} at timestep \textit{t} using:

\vspace{-1mm}
\begin{equation}
    \textit{$h_t$} = \sum_{j=1}^{k+1} \alpha_j h_j^{'}
\end{equation}

Intuitively, the \textit{reset gate} controls the information flow from the multiple predecessors to the hidden state of current timestep. If a precedent word is more correlated to the current input word, then it is expected to let the information of the precedent word flow through the gate to affect the representation of current timestep.

\subsubsection{Bi-directional Representation}
To obtain a bi-directional representation for the dialogue history, we use the same cell architecture (Section \ref{graphencodercell}) to loop over the \textit{forward graph} and \textit{backward graph} separately, and compute a forward representation $\stackrel{\rightarrow}{h_n}$ and a backward representation $\stackrel{\leftarrow}{h_n}$, respectively. Then we concatenate them together to serve as the final representation of dialogue history \textit{$h_n^{e}$}=[$\stackrel{\rightarrow}{h_n}$;$\stackrel{\leftarrow}{h_n}$], which will become a part of the inputs to the decoder.

\subsection{Multi-hop Reasoning Mechanism over Knowledge Graph}\label{graphknowledge}
A straightforward way to explore the graph information in KB is to represent the KB as a graph structure, and then query the graph using attention mechanism with the decoder hidden states. However, our preliminary experiments didn't show a good performance using this approach. We conjecture that it may be due to the poor reasoning ability of this method. To address this issue, we extend the graph with multi-hop reasoning mechanism, which aimed to strengthen the reasoning ability over graph as well as to capture the graph structural information between entities via self-attention. We call it knowledge graph module in the following sections.

Formally, the knowledge graph module contains two sets of trainable parameters \textit{C} = \{\textit{$C^{1}$},\textit{$C^{2}$},\dots,\textit{$C^{K+1}$}\}, where each \textit{$C^{k}$} is an embedding matrix that maps tokens to vector representations, and \textit{V} = \{\textit{$V^{1}$},\textit{$V^{2}$},\dots,\textit{$V^{K+1}$}\}, where each \textit{$V^{k}$} is a weight vector for computing self-attention coefficients, and \textit{K} is the maximum number of hops.

Now we describe how to compute the output vector of the knowledge graph. The model loops over \textit{K} hops on an input graph. At each hop \textit{k}, a query vector \textit{$q^{k}$} is employed as the reading head. First, the model uses an embedding layer \textit{$C^{k}$} to obtain the continuous vector representations of each node \textit{i} in the graph as \textit{$C^{k}_{i}$}, where \textit{$C^{k}_{i}$}=\textit{$C^{k}$}(\textit{$n_i$}) and $n_i$ is the \textit{i}-th node in the graph. Then we perform \textit{self-attention mechanism} on the nodes and compute the attention coefficients using:

\vspace{-1mm}
\begin{equation}
    \textit{$e_{ij}$} = \varphi\left(\left(\textit{{$V^{k}$}}\right)^{T}[\textit{$C^{k}_{i}$}||\textit{$C^{k}_{j}$}]\right) \label{gatlogits}
\end{equation}

\noindent where $\varphi$ is the LeakyReLU activation function (with negative input slope $\alpha$ = 0.2), \textit{$V^{k}$} is the parametrized weight vector of the attention mechanism at hop \textit{k}, \textit{$C^{k}_{i}$} and \textit{$C^{k}_{j}$} are the node vectors for the \textit{i}-th and \textit{j}-th node in the graph at hop \textit{k}, and $\|$ is the concatenation operation. We then normalize the coefficients of each node \textit{i} with respect to all its first-order neighbors using the softmax function:

\vspace{-1mm}
\begin{equation}
    \alpha_{ij} = \frac{exp(e_{ij})}{\sum_{k\in\textit{$N_i$}}         exp(e_{ik})} \label{gatattention}
\end{equation}

\noindent where \textit{$N_i$} is the first-order neighbors of node \textit{i} (including \textit{i}), \textit{exp} is the exponential function.

Then we update the representation of each node \textit{i} by a weighted sum of its neighbors in \textit{$N_i$} using:

\vspace{-1mm}
\begin{equation}
    \left(\textit{$C^{k}_{i}$}\right)^{'} = \sum_{j\in\textit{$N_i$}} \alpha_{ij} C^{k}_{j} \label{gatupdate}
\end{equation}

\noindent Next, the query vector \textit{$q^k$} is used to attend to the updated nodes in the graph and compute the attention weights for each node \textit{i} at hop \textit{k} using:

\vspace{-1mm}
\begin{equation}
    \textit{$p^{k}_{i}$} = softmax \left(\left(q^{k}\right)^{T} \left(\textit{$C^{k}_{i}$}\right)^{'} \right)
\end{equation}

To obtain the output of the knowledge graph, we apply the same self-attention mechanism (Equations \ref{gatlogits} and \ref{gatattention}) and update strategy (Equation \ref{gatupdate}) to the node representation \textit{$C^{k+1}_{i}$}. We use \textit{$C^{k+1}$} here since the adjacent weighted tying strategy is adopted. The updated node representation for output is denoted as $\textit{$\left(\textit{$C^{k+1}_{i}$}\right)$}^{'}$. Once obtained, the model reads out the graph \textit{$o^k$} by the weighted sum over it using:

\vspace{-1mm}
\begin{equation}
    \textit{$o^{k}$} = \sum_{i} p^{k}_{i} \left(C^{k+1}_{i}\right)^{'}
\end{equation}

Then the query vector \textit{$q^k$} is updated for the next hop using \textit{$q^{k+1}$} = \textit{$q^k$} + \textit{$o^k$}. The final output of the knowledge graph is \textit{$o^{K}$}, which will become a part of the inputs to the decoder.

\subsubsection{Graph Construction}\label{graphcontent}
In practice, dialogue systems usually use KBs (mostly in a relational database format) to provide external knowledge. We have converted the original relational database into a graph structure to exploit the relation information between KB entities. First, we find all the entities in the relational database as the nodes of the graph. Then we assign an edge to a pair of entities if there exists relationship between them according to the records in the relational database. Thus we can obtain the graph structured external knowledge.

\subsection{Decoder}\label{graphdecoder}
We use a standard Gated Recurrent Unit (GRU) \citep{cho2014learning} as the decoder to generate the system response word-by-word. The initial hidden state \textit{$h_0$} consists of two parts: the graph encoder output and the knowledge graph output. We take the output hidden state of the graph encoder \textit{$h_{n}^{e}$} as the initial query vector \textit{$q^{0}$} to attend to the knowledge graph and obtain the output \textit{$o^{K}$}. The initial hidden state \textit{$h_0$} is then computed using:

\vspace{-1mm}
\begin{equation}
    \textit{$h_0$} = [h_{n}^{e} || o^{K}]
\end{equation}

At each decoder timestep \textit{t}, the GRU takes the previously generated word \textit{$\widehat{y}_{t-1}$} and the previous hidden state \textit{$h_{t-1}$} as the input and generates a new hidden state \textit{$h_{t}$} using:
 
\vspace{-1mm}
 \begin{equation}
    \textit{$h_t$} = GRU\left(\widehat{y}_{t-1},h_{t-1} \right)
\end{equation}

Next, we follow \citep{wu2019global} that the decoder learns to generate a sketch response that the entities in the response are replaced with certain tags. The tags are obtained from the provided ontologies in the training data. The hidden state \textit{$h_t$} are used for two purposes. The first one is to generate a vocabulary distribution \textit{$P_{vocab}$} over all the words in the vocabulary using:

\vspace{-1mm}
\begin{equation}
    \textit{$P_{vocab}$} = softmax\left(W_o h_t \right)
\end{equation}

\noindent where $W_o$ is the learnable parameter. The second one is to query the knowledge graph to generate a graph distribution \textit{$P_{graph}$} over all the nodes in the graph. We use the attention weights at the last hop of the knowledge graph \textit{$p^{K}_{t}$} as \textit{$P_{graph}$}.

At each timestep \textit{t}, if the generated word from \textit{$P_{vocab}$} (the word has the maximum posterior probability) is a tag, then the decoder choose to copy from the graph entities that has the largest attention value according to \textit{$P_{graph}$}. Otherwise, the decoder will generate the target word from \textit{$P_{vocab}$}. During training, all the parameters are jointly learned via minimizing the sum of two cross-entropy losses: one is between \textit{$P_{vocab}$} and \textit{$y_t\in$} \textit{Y}, and the other is between \textit{$P_{graph}$} and \textit{$G_{t}^{Label}$}, where \textit{$G_{t}^{Label}$} is the node id that corresponds to the current output \textit{$y_t$}.

\section{Experiments}
\subsection{Dataset}
To validate the efficacy of our proposed model, we evaluate it on two public multi-turn task-oriented diaglogue datasets: Stanford multi-domain dialogue (SMD) \citep{eric2017key} and MultiWOZ 2.1 \citep{eric2019multiwoz}. The SMD is a human–human dataset for in-car navigation task. It includes three distinct task domains: \textit{point-of-interest navigation}, \textit{calendar scheduling} and \textit{weather information retrieval}. The MultiWOZ 2.1 dataset is a recently released human–human dialogue corpus with much larger data size and richer linguistic expressions that make it a more challenging benchmark for end-to-end task-oriented dialogue modeling. It consists of seven distinct task domains: \textit{restaurant}, \textit{hotel}, \textit{attraction}, \textit{train}, \textit{hospital}, \textit{taxi} and \textit{police}. We select four domains (\textit{restaurant}, \textit{hotel}, \textit{attraction}, \textit{train}) to test our model since the other three domains (\textit{police}, \textit{taxi}, \textit{hospital}) lack KB information which is essential to our task. We will make our code and data publicly available for further study. To the best of our knowledge, we are the first to evaluate end-to-end task-oriented dialogue models on MultiWOZ 2.1. The train/validation/test sets of these two datasets are split in advance by the providers.

\begin{table}[t]
    \centering
    \small
    \begin{tabular}{l|c|c}
         \toprule[1pt]
         Metrics& SMD& MultiWOZ 2.1  \\
         \midrule[0.5pt]
         \textit{Avg. Turns per dialog}& 5.25& 13.46 \\
         \textit{Avg. Tokens per turn}& 8.02& 13.13 \\
         \textit{Total number of turns}& 12732& 113556 \\
         \midrule[0.5pt]
         \textit{Vocabulary}& 1601& 23689 \\
         \textit{Train dialogs}& 2425& 8438 \\
         \textit{Val dialogs}& 302& 1000 \\
         \textit{Test dialogs}& 304& 1000 \\
         \bottomrule[1pt]
    \end{tabular}
    \caption{Dataset statistics for SMD and MultiWOZ 2.1.}
    \label{datasetstatistics}
\end{table}

\begin{table*}[!h]
    \centering
    \small
    \begin{tabular}{r|ccccccc|ccc}
        \toprule[1pt]
         \multirow{2}{*}{Model}& \multirow{2}{*}{S2S} & \multirow{2}{*}{S2S + Attn} & \multirow{2}{*}{Ptr-Unk} & \multirow{2}{*}{GraphLSTM} & \multirow{2}{*}{BERT} & \multirow{2}{*}{Mem2Seq} & \multirow{2}{*}{GLMP} & \multicolumn{3}{c}{GraphDialog} \\
         \cline{9-11}
         &  &   &  &  &  &  &  & K=1 & K=3 & K=6 \\
         \midrule[0.5pt]
         BLEU&  8.4&   9.3&  8.3&  10.3&  9.13&  12.6&  12.2&  12.96& \textbf{13.66}& 12.74 \\
         \midrule[0.5pt]
         Entity F1&  10.3&   19.9&  22.7&  50.8&  49.6&  33.4&  55.1&  56.14& \textbf{57.42}& 55.90 \\
         \midrule[0.5pt]
         Schedule F1&  9.7&  23.4 &  26.9&  69.9&  57.4&  49.3&  67.3&  70.96& \textbf{71.90}& 71.84 \\
         Weather F1&  14.1&  25.6&  26.7& 46.6&  47.5&  32.8&  54.1& 56.89& \textbf{59.68}& 54.36 \\
         Navigation F1&  7.0&  10.8&  14.9&  43.2&  46.8&  20.0&  48.4&  48.37& \textbf{48.58}& 47.55 \\
        \bottomrule[1pt]
    \end{tabular}
    \caption{Evaluation on SMD dataset. Human, rule-based and KV Retrieval Net results are reported from \citep{eric2017key}, which are not directly comparable since the problem is simplified to canonicalized forms. K denotes the maximum number of hops for knowledge graph. Ours achieves highest BLEU and entity F1 score over baselines.}
    \label{smdresults}
\end{table*}

\begin{table*}[!h]
\setlength{\belowcaptionskip}{-1mm}
    \centering
    \small
    \begin{tabular}{r|ccccccc|ccc}
         \toprule[1pt]
         \multirow{2}{*}{Model}& \multirow{2}{*}{S2S} & \multirow{2}{*}{S2S + Attn} & \multirow{2}{*}{Ptr-Unk} & \multirow{2}{*}{GraphLSTM} & \multirow{2}{*}{BERT} & \multirow{2}{*}{Mem2Seq} & \multirow{2}{*}{GLMP} & \multicolumn{3}{c}{GraphDialog} \\
         \cline{9-11}
         &  &   &  &  &  &  &  & K=1 & K=3 & K=6 \\
         \midrule[0.5pt]
         BLEU&  2.5&   3.0&  2.3&  3.4&  3.9&  4.1&  4.3&  5.47& \textbf{6.17}& 5.14 \\
         \midrule[0.5pt]
         Entity F1&  1.3&  2.1&  2.5&  4.7& 4.1&  3.2&  6.7&  9.56&  \textbf{11.28}& 8.74 \\
         \midrule[0.5pt]
         Restaurant F1&  1.6&   2.2&  2.3& 9.8&  7.3&  2.9&  11.4&  15.27& \textbf{15.95}& 13.25 \\
         Hotel F1&  1.5&  3.4&  3.8& 2.1&  1.6&  4.5&  3.9& 7.54& \textbf{10.79}& 7.05 \\
         Attraction F1&  0.8&  1.4&  1.7&  7.2&  8.4&  2.1&  9.4&  5.78& \textbf{14.12}& 7.89 \\
         Travel F1&  0.2&  0.7&  0.9&  1.8&  2.1&  1.5&  3.5&  3.41&  \textbf{4.39}& 3.53 \\
         \bottomrule[1pt]
    \end{tabular}
    \caption{Evaluation on MultiWOZ 2.1 dataset. Ours achieves highest BLEU and entity F1 score over baselines.}
    \label{multiwozresults}
\end{table*}

\subsection{Training Details}
We implement our model\footnote{Code and data are available at: \url{https://github.com/shiquanyang/GraphDialog}} in Tensorflow and is trained on NVIDIA GeForce RTX 2080 Ti. We use grid search to find the best hyper-parameters for our model over the validation set (use BLEU as criterion for both datasets). We randomly initialize all the embeddings in our implementation. The embedding size is selected between [16,512], which is also equivalent to the RNN hidden state (including the encoder and the decoder). We also use dropout for regularization on both the encoder and the decoder to avoid over-fitting and the dropout rate is set between [0.1,0.5]. We use Adam optimizer \citep{kingma2014adam} to accelerate the convergence with a learning rate chosen between [\textit{$1e^{-3}$},\textit{$1e^{-4}$}]. We simply use a greedy strategy to search for the target word in the decoder without advanced techniques like beam-search.

\subsection{Evaluation Metrics}
We use two common evaluation metrics in dialogue studies including BLEU \citep{papineni2002bleu} (using Moses \verb|multi-bleu.perl| script) and Entity F1 \citep{eric2017key,mem2seq} for evaluations.

\subsection{Effect of Models}
We compare our model with several existing models: standard sequence-to-sequence (Seq2Seq) models with and without attention \citep{luong2015effective}, pointer to unknown (Ptr-Unk, \citep{gulcehre2016pointingunknownwords}), GraphLSTM \citep{peng2017cross}, BERT \citep{devlin2019bert}, Mem2Seq \citep{mem2seq} and GLMP \citep{wu2019global}. Note that the results we listed in Table \ref{smdresults} for GLMP is different from the original paper, since we re-implement their model in Tensorflow according to their released Pytorch code for fair comparison.

\textbf{Stanford Multi-domain Dialogue}. Table \ref{smdresults} has shown the results on SMD dataset. Our proposed model achieves a consistent improvement over all the baselines with the highest BLEU score 13.6 and 57.4\% entity F1 score. The performance gain in BLEU score suggests that the generation error in the decoder has been reduced. The improvement on entity F1 indicates that our model can retrieve entities from the external knowledge data more accurately than those baselines. We also conduct comparisons with BERT to validate the effectiveness of our proposed model. Specifically, we use the bert-base-uncased model (due to GPU memory limit) from huggingface library\footnote{\url{https://github.com/huggingface}} as our encoder to encode the dialogue history and the remaining parts are the same as our model. We then fine-tune BERT on our dialogue dataset. We can find that our mode significantly outperforms the fine-tuned BERT by a large margin which further demonstrates the effectiveness of our proposed model. We conjecture that the reasons may lie in two aspects. First, the context of the corpus used for pretraining BERT differs from our dialogue dataset. Secondly, the model complexity of BERT may cause overfitting issue on small-scale datasets like SMD etc.

\textbf{MultiWOZ 2.1}. Table \ref{multiwozresults} shows the results on a more complex dataset MultiWOZ 2.1. Our model outperforms all the other baselines by a large margin both in entity F1 and BLEU score, which confirms our model has a better generalization ability than those baselines. One may find that the entity F1 and BLEU score has a huge gap between MultiWOZ 2.1 and SMD. This performance degradation phenomenon has also been observed by other dialogue works \citep{budzianowski2018multiwoz} which implies that the MultiWOZ corpus is much more challenging than the SMD dataset for dialogue tasks.

\begin{figure*}[!h]
    \centering
    \subfigure[Generation \textit{timestep 0}]{\includegraphics[width=1.5in]{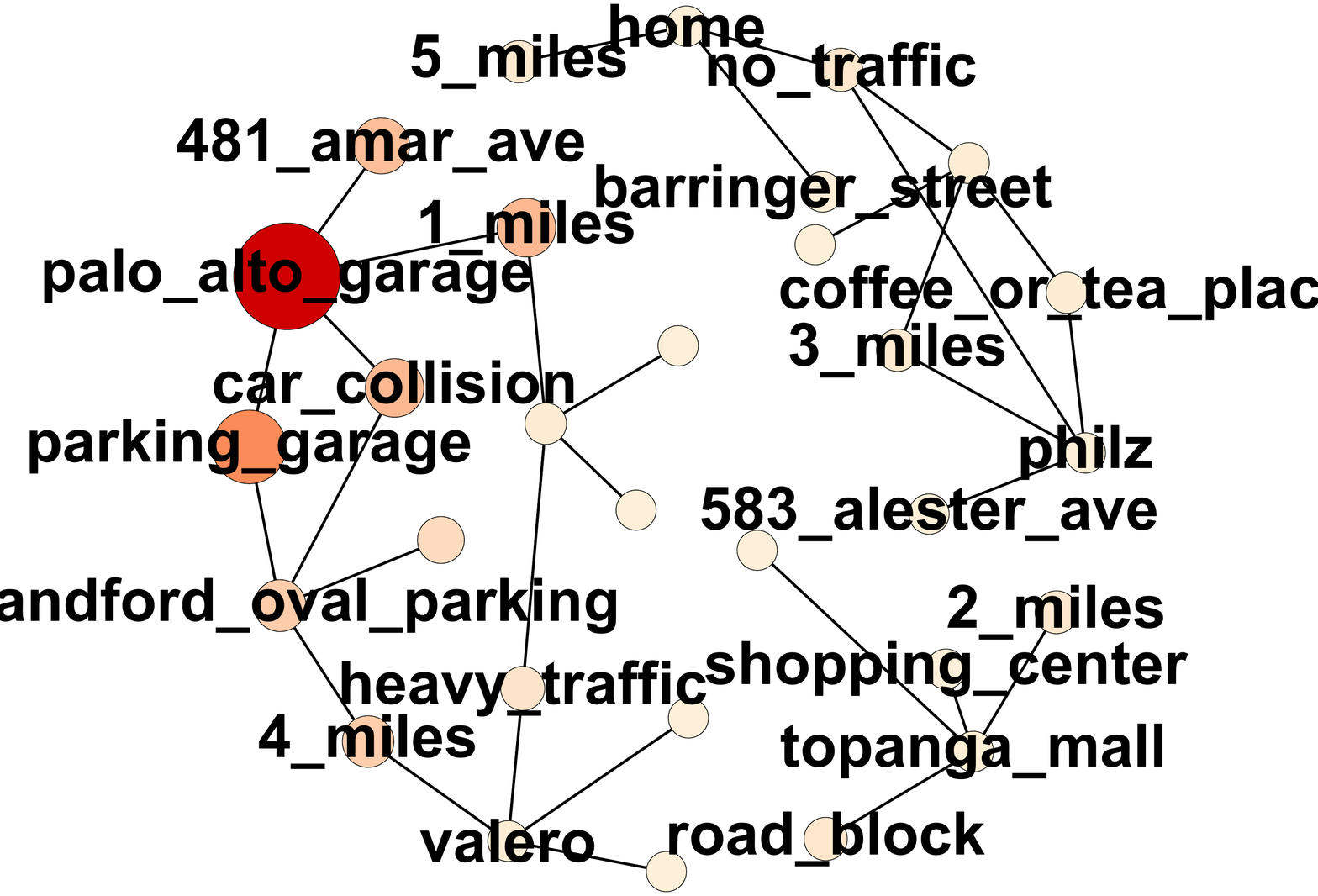}}
    \subfigure[Generation \textit{timestep 1} ]{\includegraphics[width=1.5in]{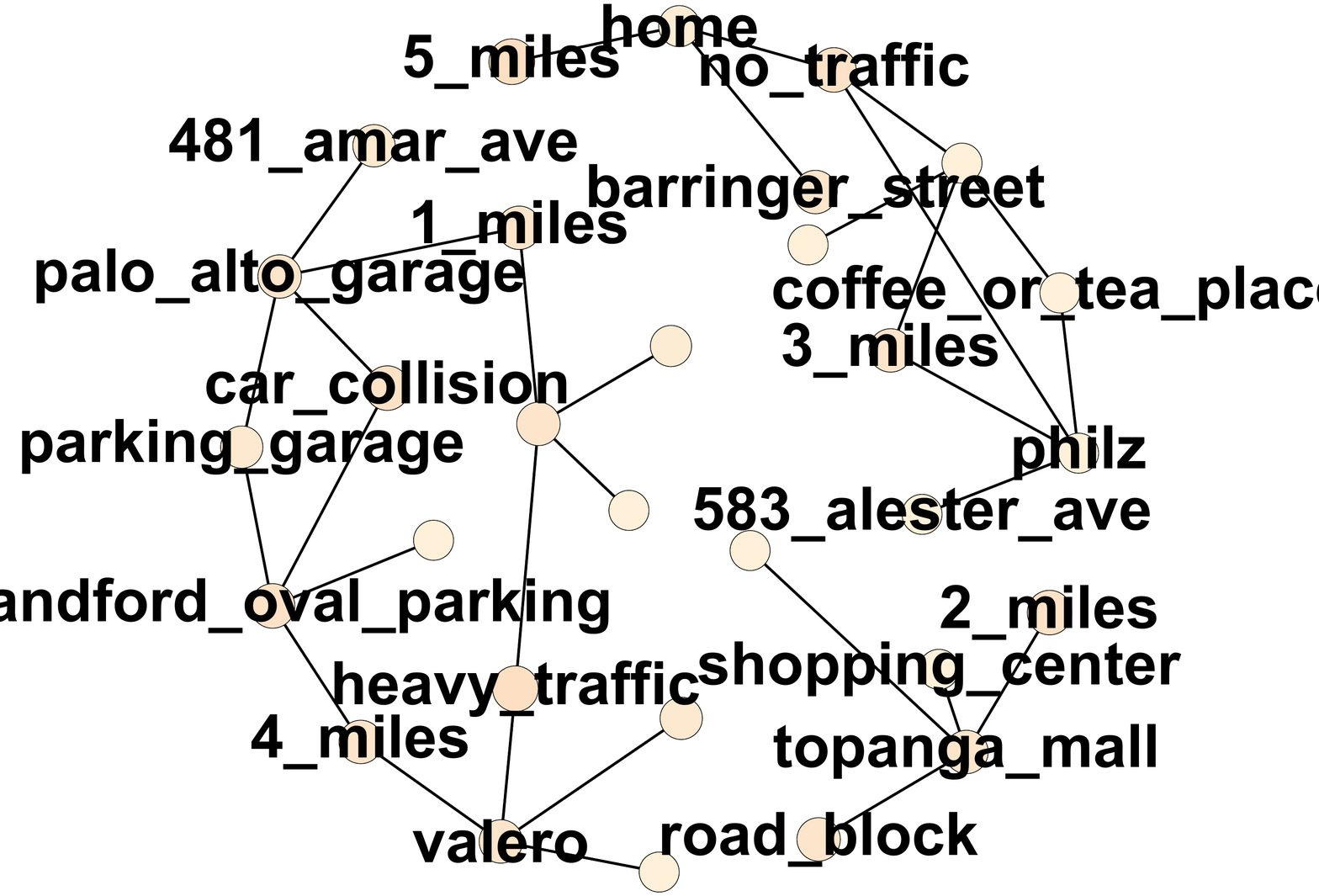}}
    \subfigure[Generation \textit{timestep 2}]{\includegraphics[width=1.5in]{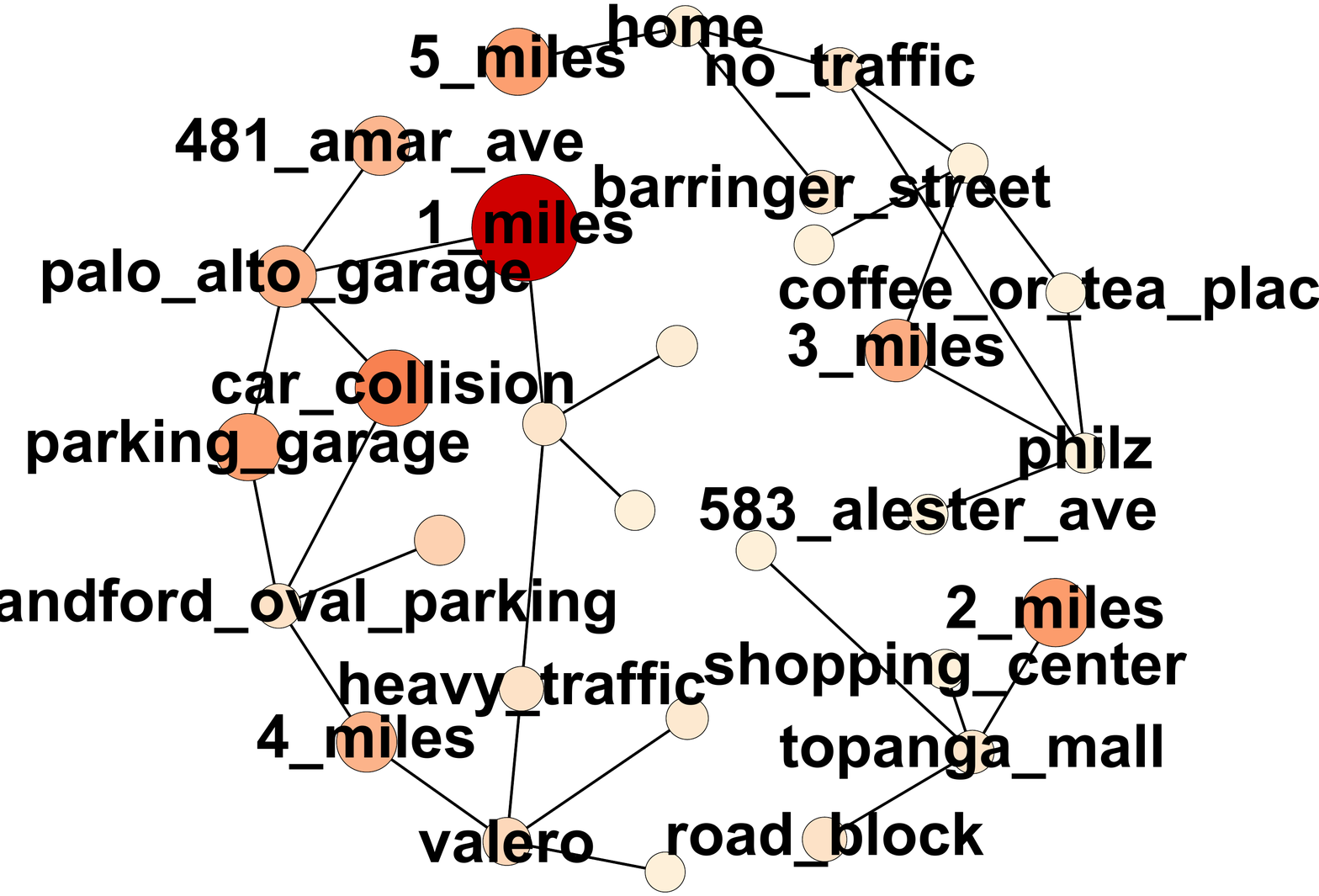}}
    \subfigure[Generation \textit{timestep 3}]{\includegraphics[width=1.5in]{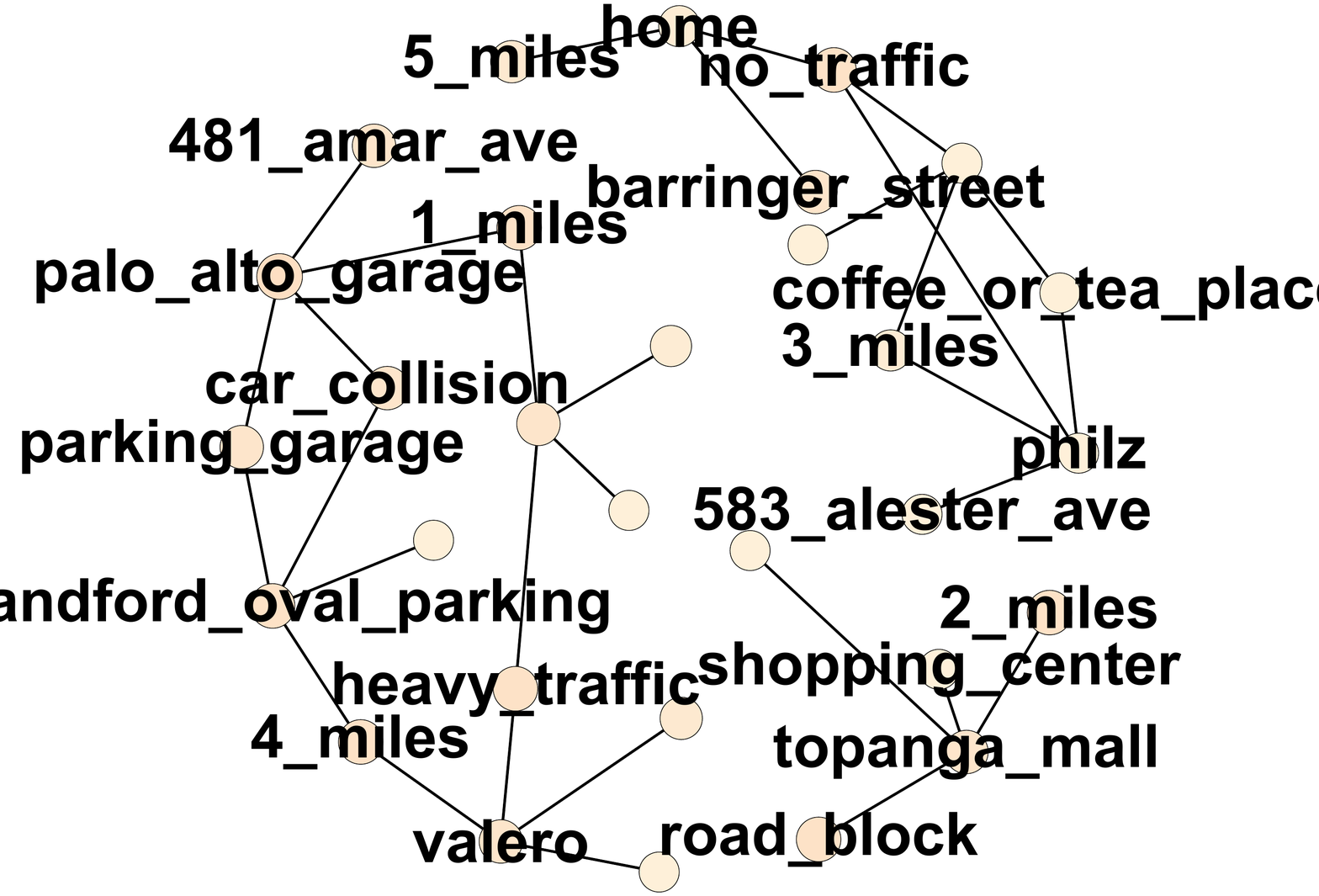}}
    \caption{Knowledge graph attention visualization when generating responses in the SMD navigation domain. Based on the question ``Where is a nearby parking\_garage?", the generated response of our model is ``palo\_alto\_garage is 1\_miles away". Specifically, the attention results at each generation timestep for the knowledge graph information of this example are shown in (a), (b), (c) and (d) respectively. The color and size of the nodes represent their attention weights. The darker and bigger the nodes are, the larger their attention weights are. Our model successfully learns to attend to the correct KB entities (i.e., \textit{palo\_alto\_garage} and \textit{1\_miles} at generation timesteps 0 and 2) which have the highest attention, and the model copies them to serve as the output words. During timesteps 1 and 3, the model generates output words (i.e., \textit{is} and \textit{away}) from the vocabulary.}
    \label{kgattention}
 \end{figure*}

\begin{table*}[h]
    \centering
    \small
    \begin{tabular}{l|cccc}
         \toprule[1pt]
         & \multicolumn{2}{c|}{SMD}& \multicolumn{2}{c}{MultiWOZ 2.1} \\
         \midrule[0.5pt]
         Model& BLEU& \multicolumn{1}{c|}{Entity F1(All)}& BLEU& Entity F1(All) \\
         \midrule[0.5pt]
         GraphDialog & 13.66(-)& \multicolumn{1}{c|}{57.42(-)} & 6.17(-)& 11.28(-) \\
         GraphDialog w/o Graph Encoder & 12.35(-1.31)& \multicolumn{1}{c|}{56.61(-0.81)}& 4.57(-1.60)& 10.13(-1.15)  \\
         GraphDialog w/o Knowledge Graph & 13.13(-0.53)&  \multicolumn{1}{c|}{55.28(-2.14)}& 5.35(-0.82)& 7.41(-3.87)  \\
         \bottomrule[1pt]
    \end{tabular}
    \caption{Model ablation study: Effects of Graph Encoder and Knowledge Graph. Number in the parentheses means the absolute value gap between the full version and the ablation one on corresponding metrics.}
    \label{ablationstudy}
\end{table*}

\textbf{Ablation Study.} Table \ref{ablationstudy} shows the contributions of each components in our model. Ours without graph encoder means that we do not use the dependency relations information and the proposed recurrent cell architecture. We simply use a bi-directional GRU to serve as the encoder and the other parts of the model remain unchanged. We can observe that our model without the graph encoder has a 1.6\% absolute value loss (over 25\% in ratio) in BLEU score and a 1.1\% absolute value loss (9.8\% in ratio) in entity F1 on MultiWOZ 2.1, which suggests that the overall quality of the generated sentences are better improved by our graph encoder. On the other hand, ours without knowledge graph means that we do not use the graph structure to store and retrieve the external knowledge data. Instead we use memory networks \citep{sukhbaatar2015end} that has been shown useful to handle the knowledge base similar to \citep{wu2019global}. We can find a significant entity F1 drop (3.8\% in absolute value and 33.9\% in ratio) on MultiWOZ 2.1, which verifies the superiority of the proposed graph-based module with multi-hop reasoning ability in retrieving the correct entities, even compared to the strong memory-based baselines.

\textbf{Model Training Time.} We also compare the training time of GraphDialog with those baselines. GraphDialog is about 3 times faster than BERT since its model complexity is smaller. The number of parameters for GraphDialog is almost 90\% less than BERT, which also saves space for model storage. GraphDialog is slower than GLMP, which is expected as it needs to encode more information. However, the gap of the training time is up to 69\%, and we can complete the whole training process within one day which seems reasonable.

\begin{table}[!t]
\setlength{\belowcaptionskip}{-4mm}
    \centering
    \small
    \begin{tabular}{c|c|c|c|c}
         \toprule[1pt]
         & \multicolumn{4}{c}{Edge Path Distance} \\
         \midrule[0.5pt]
         Dataset& 1& $\geq$ 2& $\geq$ 10& $\geq$ 15 \\
         \midrule[0.5pt]
         SMD& 52.82\%& 33.68\%& 10.61\%&2.89\%  \\
         \midrule[0.5pt]
         MultiWOZ 2.1& 50.29\%& 35.41\%& 11.26\%&3.04\%  \\
         \bottomrule[1pt]
    \end{tabular}
    \caption{Edge path distance distribution on different datasets.}
    \label{edgepath}
\end{table}

\subsection{Analysis and Discussion}\label{analysis}

\textbf{Why does dependency relations help?} We have conducted in-depth analyses from the edge path distance perspective. Table \ref{edgepath} shows the edge path distance distribution in the dialogue graph (Section \ref{dialoguegraph}) on both SMD and MultiWOZ 2.1. The edge path distance is defined as the the number of words between the head word and the tail word along the linear word sequence plus one. For example, for the sentence ``There is a supermarket", the edge distance of the ``Next" edge between ``There" and ``is" is 1, the edge path distance of the ``nsubj" edge  between ``is" and ``supermarket" is 2. We can find that although many edges have small edge path distances, there are still a considerable number of edges with relatively large distances, which could encourage more direct information flow between distant words in the input. This may partly explain the benefits of using information such as dependency relations in encoding the dialogue history.

\textbf{Attention Visualization.} To further understand the model dynamics, we analyze the attention weights of the knowledge graph module to show its reasoning process. Figure \ref{kgattention} has shown an example of the attention distribution over all the nodes at the last hop of the knowledge graph. Based on the question ``Where is a nearby parking\_garage?" asked by the user, the generated response of our model is ``palo\_alto\_garage is 1\_miles away", and the gold answer is ``The nearest one is palo\_alto\_garage, it's just 1\_miles away". We can find that our model has successfully learned to  copy the correct entities (i.e., \textit{palo\_alto\_garage} at timestep 0 and \textit{1\_miles} at timestep 2) from the knowledge graph.

\textbf{Error Analysis.} To inspire future improvements, we also inspect the generated responses manually. We find that the model tends to omit entities when the responses contain multiple KB entities. Besides, about 10\% of the generated responses contain duplicate KB entities. For example, ``\textit{The temperature in New York on Monday is 100F, 100F}". This may be attributed to the training of GRU in the decoder, and we aim to solve the problem in future work.

\section{Conclusion}
In this work, we present a novel graph-based end-to-end model for task-oriented dialogue systems. The model leverages the graph structural information in dialogue history via the proposed recurrent cell architecture to capture the semantics of dialogue history. The model further exploits the relationships between entities in the KB to achieve better reasoning ability by combining the multi-hop reasoning ability with graph. 

We empirically show that our model outperforms the state-of-the-art models on two real-world task-oriented dialogue datasets. Our model may also be applied to end-to-end open-domain chatbots since the goal is to generate responses given inputs and external knowledge, which is what our model can do. We will explore this direction in future work.

\section*{Acknowledgements}
We would like to thank Bayu Distiawan Trisedya for his insightful discussions and the valuable feedbacks from all anonymous reviewers. This work is supported by Australian Research Council (ARC) Discovery
Project DP180102050.

\bibliographystyle{acl_natbib}
\bibliography{emnlp2020}

\appendix

\end{document}